\documentclass[10pt,twocolumn,letterpaper]{article}

\usepackage{cvpr}
\usepackage{times}
\usepackage{epsfig}
\usepackage{graphicx}
\usepackage{amsmath}
\usepackage{amssymb}
\usepackage{bm}
\usepackage[breaklinks=true,bookmarks=false]{hyperref}

\cvprfinalcopy 

\def\cvprPaperID{****} 

\begin{document}

\title{PPDM: Parallel Point Detection and Matching for Real-time Human-Object
Interaction Detection}

\author{Yue Liao\textsuperscript{\rm 1,2}\quad Si Liu\textsuperscript{\rm 1}\thanks{Corresponding author (liusi@buaa.edu.cn)} \quad Fei Wang\textsuperscript{\rm 2}\quad Yanjie Chen\textsuperscript{\rm 2}\quad Chen Qian\textsuperscript{\rm 2}\quad Jiashi Feng\textsuperscript{\rm 3}\\
\large\textsuperscript{\rm 1} School of Computer Science and Engineering, Beihang University \\\quad\textsuperscript{\rm 2} SenseTime Research\quad\textsuperscript{\rm 3} National University of Singapore}

\maketitle
\begin{abstract}
We propose a single-stage Human-Object Interaction (HOI) detection method that has outperformed all existing methods on HICO-DET dataset at $37$ fps on a single Titan XP GPU. It is the first real-time HOI detection method. 
Conventional HOI detection methods are composed of two stages, i.e., human-object proposals generation, and proposals classification. Their effectiveness and efficiency are limited by the sequential and separate architecture. 
In this paper, we propose a Parallel Point Detection and Matching (PPDM) HOI detection framework. In PPDM, an HOI is defined as a point triplet $<$ human point, interaction point, object point$>$. Human and object points are the center of the detection boxes, and the interaction point is the midpoint of the human and object points. 
PPDM contains two parallel branches, namely point detection branch and point matching branch.
The point detection branch predicts three points. Simultaneously, the point matching branch predicts two displacements from the interaction point to its corresponding human and object points. The human point and the object point originated from the same interaction point are considered as matched pairs. 
In our novel parallel architecture, the interaction points implicitly provide context and regularization for human and object detection. The isolated detection boxes unlikely to form meaningful HOI triplets are suppressed, which increases the precision of HOI detection. Moreover, the matching between human and object detection boxes is only applied around limited numbers of filtered candidate interaction points, which saves much computational cost. 
Additionally, we build a new application-oriented database named as HOI-A, which serves as a good supplement to the existing datasets\footnote{https://github.com/YueLiao/PPDM}.
\end{abstract}
\vspace{-3mm}

\begin{figure}[h]
  \centering
  \includegraphics[width=0.98\linewidth]{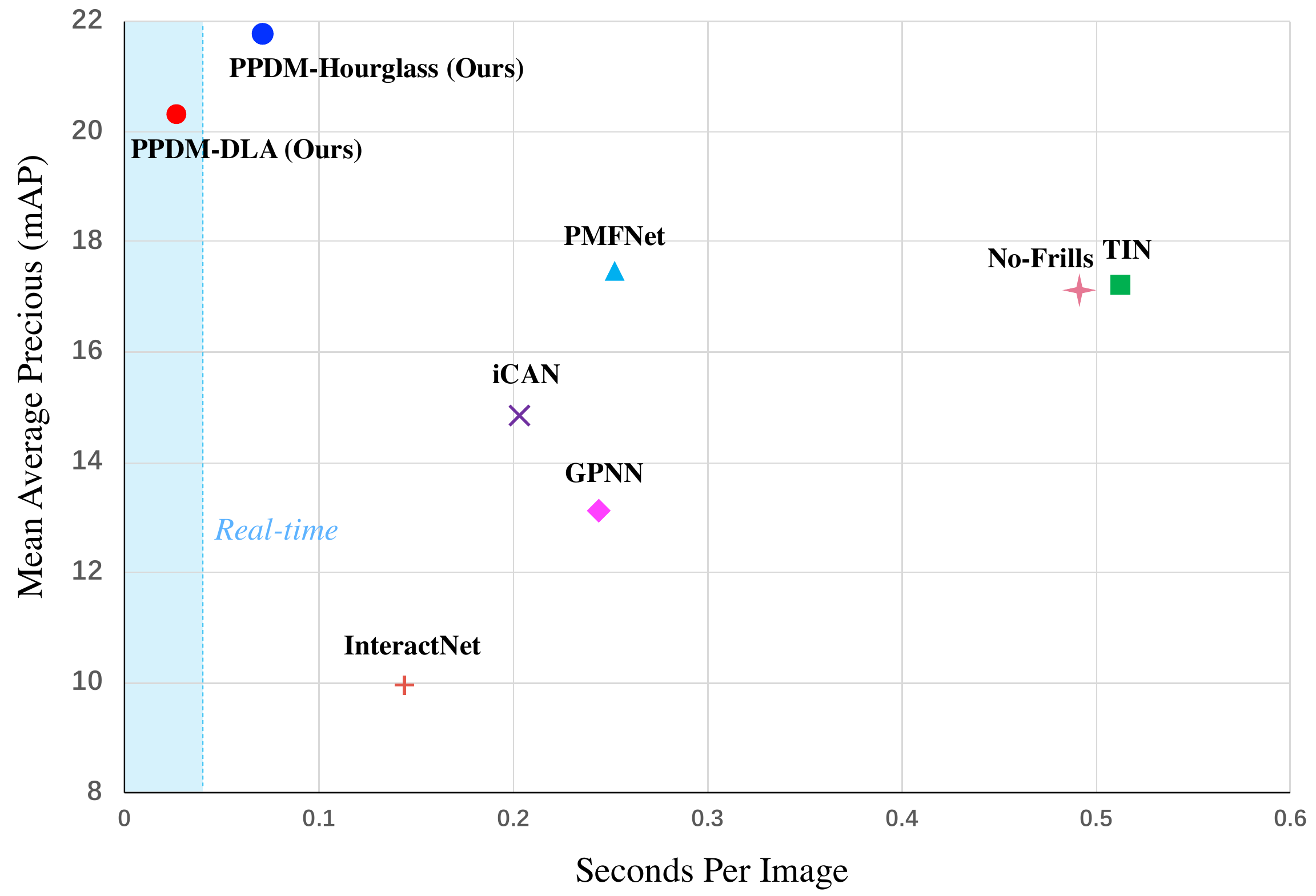}
  \vspace{-2mm}
  \caption{mAP versus inference time on the HICO-Det test set. Our PPDM-DLA outperforms the state-of-the-art methods with the inference speed of $37$ fps ($0.027$s). It is the first real-time HOI detection method. Our PPDM-Hourglass achieves $4.27\%$ mAP improvement over the state-of-the-arts with a faster speed.}
  \label{fig:first}
   \vspace{-3.5mm}
\end{figure}
\section{Introduction}
Human-Object Interaction (HOI) detection \cite{yao2012recognizing,gupta2015visual,gupta2009observing,gkioxari2018detecting,gupta2018no,li2018transferable,qi2018learning} has received increasing attention recently. Given an image, HOI detection aims to detect the triplet $<$ human, interaction, object $>$. Different from the general visual relationship detection \cite{lu2016visual,xu2017scene,newell2017pixels,herzig2018mapping,zhang2019graphical}, the subject of the triplet is fixed as human while the interaction is action.  HOI detection is an important step toward the high-level semantic understanding of human-centric scenes. It has a lot of applications, such as activity analysis, human-machine interaction and intelligent monitoring.

The conventional HOI detection methods \cite{chao2018learning,qi2018learning,gupta2018no,li2018transferable,Wan_2019_ICCV}  mostly consist of two stages. The first stage is the human-object proposal generation. A pre-trained detector~\cite{girshick2015fast,ren2015faster} is used to localize both the humans and objects. Then $M \times N$ human-object proposals are generated by pairwisely combining the filtered  $M$ human boxes and $N$ object boxes. The second stage is the proposal classification which predicts the interactions for each human-object proposal.
The limitations of the two-stage methods' effectiveness and efficiency are mainly because their two stages are \emph{sequential and separated}.  The proposal generation stage is completely based on object detection confidences. Each human/object proposal is independently generated. The possibility of combining two proposals to form a meaningful HOI triplet in the second stage is not taken into account. Therefore, the generated human-object proposals may have relatively low quality.  Moreover, in the second stage, all human-object proposals need to be linearly scanned, while only a few of them are valid. The extra computational cost is large. Therefore, we argue that the \emph{non-sequential and highly-coupled} framework is needed.

\begin{figure}[t]
	\centering
	\includegraphics[width=0.7\linewidth]{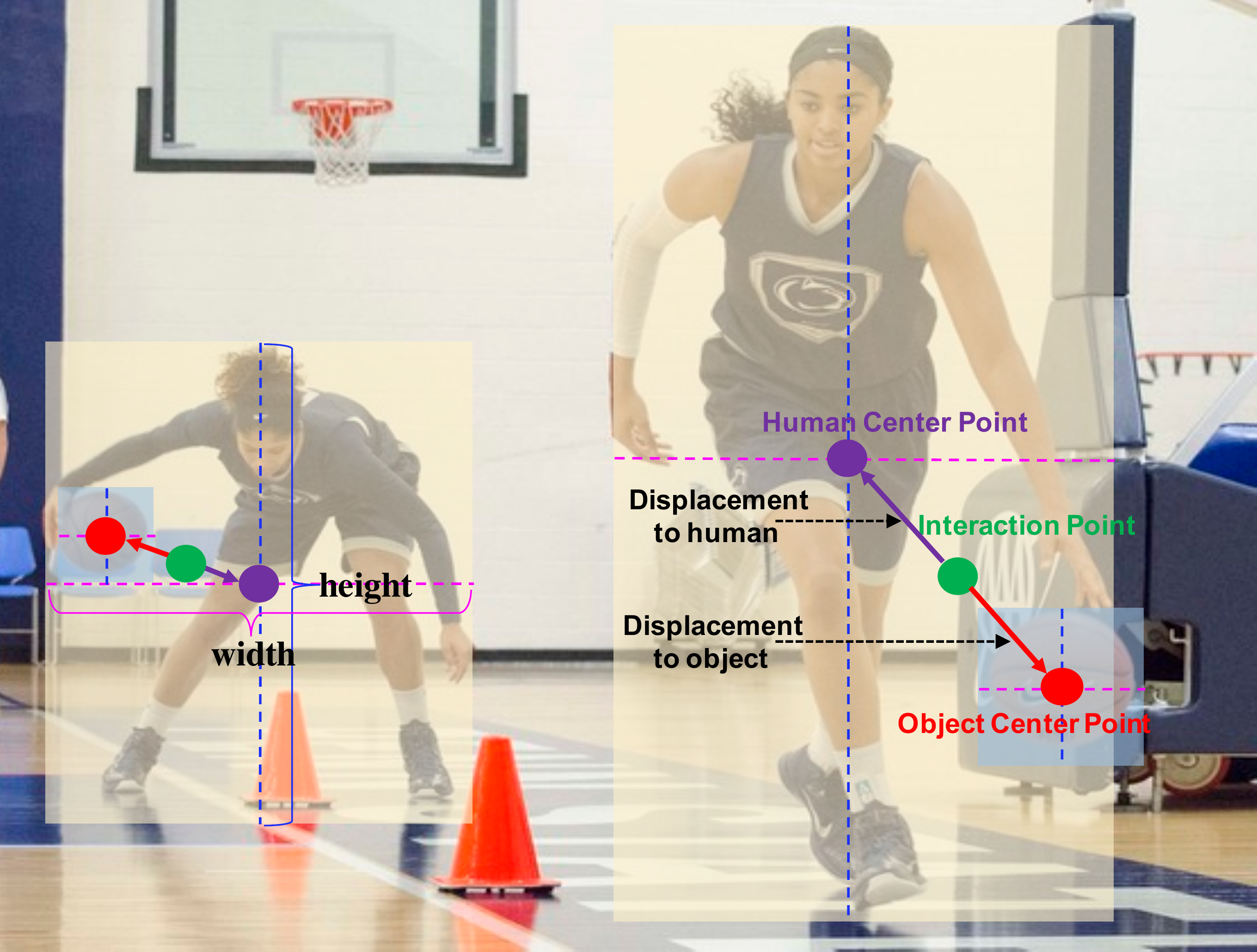}
	\caption{ PPDM contains two parallel branches. In the point detection branch, the human/object box denoted as the center points, widths, and heights, are detected. Moreover, an interaction point, i.e., the midpoint of the human and object point, is also localized.
    Simultaneously, in the point matching branch, two displacements from each interaction point to the human/object are estimated. The human point and the object point originated from the same interaction point are considered as matched pairs. }
	\label{fig:key_idea}
	  \vspace{-3mm}
\end{figure}

We propose a \emph{parallel} HOI detection framework and reformulate HOI detection as a point detection and matching problem. As shown in Figure~\ref{fig:key_idea}, we represent a box as a center point and corresponding sizes (width and height). Moreover, we define an interaction point as the midpoint of the human and object center points. To match each interaction point with the human point and the object point, we design two displacements from the interaction point to the corresponding human and object point. Based on the novel reformulation, we design a novel single-stage framework Parallel Point Detection and Matching (PPDM), which breaks up the complex task of  HOI detection into two simpler parallel tasks. The PPDM is composed of two parallel branches. The first branch is \emph{points detection}, which estimates the three center points (interaction, human and object points), corresponding sizes (width and height) and two local offsets (human and object points). 
The interaction point can be considered as providing contextual information for both human and object detection. In other words, estimating the interaction point implicitly enhances the detection of humans and objects. 
The second branch is  \emph{points matching}. Two displacements from the interaction point to human and object points are estimated.  The human and object points originated from the same interaction points are considered as matched. In the novel parallel architecture, the point detection branch estimates the interaction points, which implicitly provide context and regularization for the human and object detection. The isolated detection boxes unlikely to form meaningful HOI triplets are suppressed while the more likely detection boxes are enhanced. It is different from the human-object proposal generation stage in two-stage methods, where all detection human/object boxes indiscriminately form the human-object proposals to feed into the second stage. Moreover, in the point matching branch, the matching is only applied around limited numbers of filtered candidate interaction points, which saves a lot of computational costs. On the contrary, in the proposal classification stage of two-stage methods, all human-object proposals need to be classified. Experimental results on the public benchmark HICO-Det~\cite{chao2018learning} and our newly collected HOI-A dataset show that PPDM outperforms state-of-the-art methods in terms of accuracy and speed. 

\begin{figure*}[htb]
  \vspace{-2mm}
  \centering
  \includegraphics[width=1\linewidth]{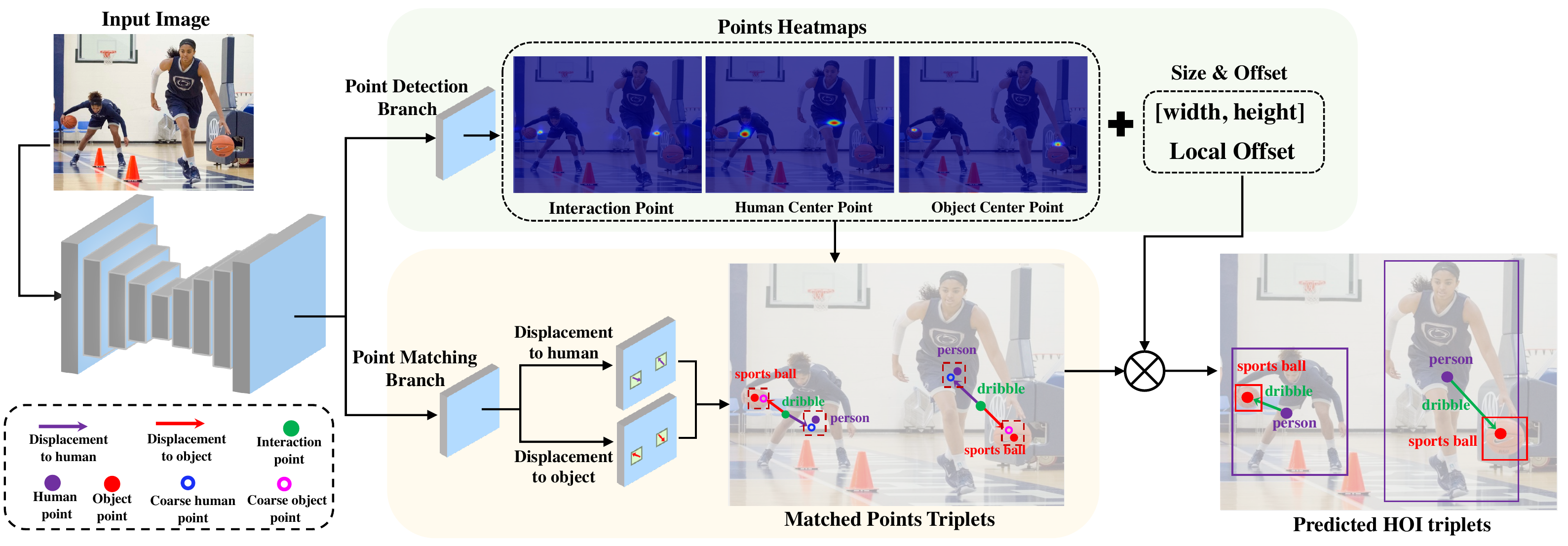}
  \vspace{-0.5mm}
  \caption{Overview of the proposed PPDM framework. We firstly apply a key-point heatmap prediction network, e.g. Hourglass-104 or DLA-34, to extract the appearance feature from an image. a) Point Detection Branch: Based on the extracted visual feature, we utilize three convolutional modules to predict the heatmap of the interaction points, human center points, and object center points. Additionally, to generate the final box, we regress the 2-D size and the local offset. b) Point Matching Branch: the first step of this branch is to regress the displacements from the interaction point to the human point and object point respectively. Based on the predicted points and displacements, the second step is to match each interaction point with the human point and object point to generate a set of points triplets. }
  \label{f_work}
  \vspace{-3mm}
\end{figure*}

The existing  datasets such as HICO-Det~\cite{qi2018learning} and V-COCO~\cite{gupta2015visual} have greatly boosted the development of related research. 
These datasets are very general.   However, in practical applications, several \emph{limited, frequent} HOI categories need to be paid special attention to. To this end,  we collect a new Human-Object Interaction for Applications dataset  (HOI-A)  with the following features: 1) specially selected $10$ kinds of HOI categories with wide application values, such as smoke and ride. 
 2) huge intra-class variations including various illuminations and different human poses for each category.  The HOI-A is more application-driven, severing as a good supplement to the existing datasets.

Our contributions are summarized as: 1) We reformulate the HOI detection task as a point detection and matching problem and propose a novel one-stage PPDM solution.  2) PPDM is the first HOI detection method to achieve real-time and outperform state-of-the-art methods both HICO-Det and HOI-A benchmarks. 3) A large-scale and application-oriented  HOI detection dataset is collected to supplement existing datasets. Both source code and the dataset are to be released to facilitate the related research.

\section{Related Work}
\noindent\textbf{HOI Detection Methods.}
The existing HOI detection methods can be mostly divided into two stages: in the first stage, an object detector~\cite{ren2015faster} is applied to localize the human and objects; in the second stage, pairing the detected human and object, and feeding their features into a classification network to predict the interaction between the human and object. Current works pay more attention to exploring how to improve the second stage. The most recent works aim to understand HOI by capturing context information~\cite{gao2018ican,Wang_2019_ICCV} or human structural message~\cite{Wan_2019_ICCV,feng2019turbo,fang2018pairwise,Zhou_2019_ICCV}. Some works~\cite{qi2018learning,Xu_2019_CVPR,Zhou_2019_ICCV} formulated the second stage as a graph reasoning problem and use graph convolutional network to predict the HOI. 

The above methods are all proposal based, thus their performance is limited by the quality of proposals. Additionally, the existing methods have to spend much computational cost in proposals generation and feature extraction process.  
Based on these drawbacks, we propose a novel one-stage and proposal-free framework to detect HOI.

\noindent\textbf{HOI Detection Datasets.}
There are mainly two commonly used HOI detection benchmarks:  VCOCO~\cite{gupta2015visual} and HICO-Det~\cite{chao2018learning}, and a human-centric relationship detection dataset: HCVRD~\cite{zhuang2017care}. The VCOCO is a relatively small dataset, which is a subset of MSCOCO~\cite{lin2014microsoft} including $10,346$ images annotated with $26$ actions based on COCO annotation. The HICO-Det is a large-scale and generic HOI detection dataset, including $47,776$ images, which has $117$ verbs and $80$ object categories (same as COCO). The HCVRD is collected from the general visual relationship detection dataset, Visual Genome~\cite{krishna2017visual}. It has $52,855$ images, $927$ predicate categories and $1,824$ kinds of objects.  Comparing the former two HOI detection datasets, which only focuses on human actions,
 the HCVRD is concerned about a more general human-centric relationship, e.g., spatial relationships, possessive relationships.

The previous HOI detection datasets mostly concentrate on common and general actions. From a practical view, we build up a new HOI-A dataset, which has about 38K images only annotated with limited typical kinds of actions with practical significance. 

\section{Parallel Point Detection and Matching}
\subsection{Overview}

HOI detection aims to estimate the HOI triplet $<$ human, interaction, object $>$, which is composed of the subject box and class, the human action class and the object box and class. 
 We break up the complex task of  HOI detection into two simpler parallel tasks that can be assembled to form the final results.
 The framework of the proposed Parallel Point Detection and Matching (PPDM) method is shown in Figure~\ref{f_work}.
The first branch of PPDM is \emph{points detection}. It estimates the center points, corresponding sizes (width and height) and local offsets of both humans and objects.
The center, size and offset collaboratively represent some box candidates.
Moreover, the interaction point which is defined as the midpoint of a corresponding  $<$ human center point, object center point $>$  pair is also estimated.  The second branch of  PPDM is  \emph{points matching}.
The displacements between the interaction point and the corresponding human and object points are estimated. The human point and the object point originated from the same interaction point are considered as matched pairs.


\subsection{Point Detection}

The point detection branch estimates the human box, object box and interaction point. A human box  is represented as its center point $(x^{h}, y^{h})\in \mathbb{R}^2$, the corresponding size (width and height)   $(w^{h}, h^{h})\in \mathbb{R}^2$ as well as  the local point offset  $\bm{\delta{c}^{h}}\in\mathbb{R}^2$   to recover the discretization error caused by the output stride. The object box is represented similarly.  Moreover, we define the interaction point $(x^{a}, y^{a})\in \mathbb{R}^2$ as the midpoint of the paired human point and object  point. Considering the receptive field of the interaction point is large enough to contain both human and object, the human action $a$ can be estimated based on the feature  of $(x^{a}, y^{a})$. Actually, when there are $M$ human in the dataset, each human box is represented as $(x_i^{h}, y_i^{h}), i \in [1,M]$. For the convenience of description, we omit the subscript $i$ when no confusion is caused. Similar omissions are also  applicable for $(x^{o}, y^{o})$ and $(x^{a}, y^{a})$.

In Figure \ref{f_work}, the input image $\bm{I}\in \mathbb{R}^{{H} \times {W}}$ is fed into the feature extractor to produce the feature $\bm{V}\in \mathbb{R}^{\frac{H}{d} \times \frac{W}{d}}$, where $W$ and $H$ are the width and height of the input image and $d$ is the output stride.
The point heatmaps are of low-resolution, thus we also calculate the low-resolution center points.
Given a ground-truth human point $(x^{h}, y^{h})$,  the corresponding low-resolution point is  $(\Tilde{x}^{h},\Tilde{y}^{h}) = (\lfloor{\frac{x^{h}}{d}}\rfloor,\lfloor{\frac{y^{h}}{d}}\rfloor)$. The low-resolution ground-truth object point $(\Tilde{x}^{o},\Tilde{y}^{o})$ can be computed in the same way. Based on the low-resolution human and object points, the ground-truth interaction point can be defined as $ (\Tilde{x}^{a}, \Tilde{y}^{a}) = (\lfloor{\frac{\Tilde{x}^{h} + \Tilde{x}^{o}}{2}}\rfloor, \lfloor{\frac{\Tilde{y}^{h} + \Tilde{y}^{o}}{2}}\rfloor)$.

\textbf{Point location loss.} Directly detecting a point is difficult, thus we follow the key-point estimation methods~\cite{tompson2014joint} to splat a point into a heatmap with a Gaussian kernel. Thereby the point detection is transformed into a heatmap estimation task. 
The three  ground-truth low-resolution points $(x^{h}, y^{h})$, $(x^{o}, y^{o})$ and $(x^{a}, y^{a})$ are splatted into three Gaussian heatmaps, including human point heatmap $\bm{\Tilde{C}^{h}}\in [0,1]^{\frac{H}{d} \times \frac{W}{d}}$, object point heatmap $\bm{\Tilde{C}}^{o} \in [0,1]^{T\times \frac{H}{d} \times \frac{W}{d}}$ and interaction point heatmap $\bm{\Tilde{C}}^{a} \in [0,1]^{K\times \frac{H}{d} \times \frac{W}{d}}$, 
where $T$ is the number of object categories and $K$ is the the number of interaction classes. Note that in $\bm{\Tilde{C}^{o}}$ and $\bm{\Tilde{C}^{a}}$, 
only the channel corresponding to the specific object class and human action are non-zero.
The three heatmaps are produced by adding three  respective  convolutional blocks upon the feature map $\bm{V}$,  each of which is composed of a $3\times3$ convolutional layer with ReLU, followed by a $1\times1$ convolutional layer and a Sigmoid.

 For the three heatmaps, we all apply an element-wise focal loss~\cite{lin2017focal}. For example,  given an estimated interaction point heatmap  $\bm{\hat{{C}}^{a}}$ and the corresponding ground-truth heatmaps $\bm{\Tilde{{C}}^{a}}$, the loss function is:
\vspace{-2mm}
\begin{equation}
L_{a}= -\frac{1}{N}\sum_{k x y}\left\{\begin{array}{cc}{(1-\hat{{C}}^{a}_{k x y})^{\alpha} \log (\hat{C}^{a}_{k x y })} & {\text { if  } \Tilde{C}^a_{ k x y}=1} \\ {(1-\Tilde{C}^{a}_{k x y})^{\beta}(\hat{C}^{a}_{k x y})^{\alpha}}& {\text { otherwise }} \\ {\log (1-\hat{C}^{a}_{k x y })}, \end{array}\right.
\end{equation}
where $N$ is equal to the number of interaction points (HOI triplet) in the image,  and  $\hat{C}^a_{ k x y}$ is  the score at location $(x,y)$ for class $k$ in the predicted heatmaps $\bm{\hat{C}^{a}}$.  We set $\alpha$ as 2 and $\beta$ as 4 following the default setting in \cite{law2018cornernet,zhou2019objects,Dong_2020_CVPR}. The losses $L_p$ and $L_o$ for the human points and the object points can be computed similarly.

\textbf{Size and offset loss.} 
Besides the  center points,  the box size and the local offset for the center points are needed to form the human/object box. 
Four convolutional blocks are added to the feature map $\bm{V}$ to estimate the 2-D size and the local offset of human and object boxes respectively. Each block contains a $3\times3$ convolutional layer with ReLU and a $1\times1$ convolutional layer. 


During training, we only compute the $L1$ loss at each location of the ground truth human point $(\Tilde{x}^{h},\Tilde{y}^{h})$ and object point $(\Tilde{x}^{o},\Tilde{y}^{o})$ and ignore all other locations. We take the loss function for the local offset as an example, while the size regression loss $L_{wh}$ is defined similarly. The ground truth local offset for the  human point localized at $(\Tilde{x}^{h},\Tilde{y}^{h})$ is defined as $({\Tilde{\delta}^x_{(\Tilde{x}^{h}, \Tilde{y}^{h} )}}, \Tilde{\delta}^y_{{(\Tilde{x}^{h}, \Tilde{y}^{h} )}}) = (\frac{x^h}{d} - \Tilde{x}^{h},\frac{y^h}{d} - \Tilde{y}^{h})$. Thus the loss function $L_{off}$ is the summation of the human box loss $L_{off}^h$ and object box loss $L_{off}^o$.
\vspace{-1mm}
\begin{equation}
 L_{off}= \frac{1}{M + D} (L_{off}^h+ L_{off}^o)
\end{equation}

\vspace{-1mm}
\begin{equation}
\label{eq:loss_h}
\begin{aligned}
 L_{off}^h =&   \sum_{ (\Tilde{x}^{h},\Tilde{y}^{h}) \in \Tilde{S}^h}({|{\Tilde{\delta}^x_{(\Tilde{x}^{h}, \Tilde{y}^{h} )}} - {\hat{\delta}^x_{(\Tilde{x}^{h}, \Tilde{y}^{h} )}}|}  \\ & + {|{\Tilde{\delta}^y_{(\Tilde{x}^{h}, \Tilde{y}^{h} )}} - {\hat{\delta}^y_{(\Tilde{x}^{h}, \Tilde{y}^{h} )}}|}  ,
\end{aligned}
\end{equation}
where $\Tilde{S}^h $  and  $\Tilde{S}^h $ denote the ground-truth human and object points sets in the training set.
$M = | \Tilde{S}^h | $ and $D = | \Tilde{S}^o | $  are the number of  human points  and object points. Note that $M$ is not necessarily  equal to $D$. For example a human may correspond to multiple actions and objects. $L_{off}^o$ is defined similarly with Equation \ref{eq:loss_h}.

\subsection{Point Matching}
The points matching branch pairs the human box with its corresponding object box by using the interaction point as the bridge. 
More specifically,  the interaction point is treated as the anchor. Two displacements $\bm{{d}^{a h}}= (d^{ah}_x,d^{ah}_y)$ and $\bm{{d}^{ao}}= (d^{ao}_x,d^{ao}_y)$, i.e.,  the displacements between interaction point vs. human/box point are estimated. The coarse human point and object point are $(x^{a}, y^{a})$ plus  $\bm{{d}^{ah}}$ and $\bm{{d}^{ao}}$ respectively.


 Our proposed displacement branch is composed of two convolutional modules.  Each module consists of a $3\times3$ convolutional layer with ReLU and a $1\times1$ convolutional layer.  The size of both  subject and object displacement maps  are
 ${2\times \frac{H}{d} \times \frac{W}{d}}$. 
 
 \textbf{Displacement loss.}  To train the displacement branch, we apply $L1$ loss for each interaction point. The ground-truth displacement from the  interaction point located at $(\Tilde{x}^{a}, \Tilde{y}^{a})$
 to the corresponding human point can be computed by $({\Tilde{d}^{hx}_{(\Tilde{x}^{a}, \Tilde{y}^{a})}}, {\Tilde{d}^{hy}_{(\Tilde{x}^{a}, \Tilde{y}^{a})}}) = (\Tilde{x}^{a} - \Tilde{x}^{h}, \Tilde{y}^{a} - \Tilde{y}^{h})$. The predicted displacement  at location of $(\Tilde{x}^{a}, \Tilde{y}^{a})$ is $({\hat{d}^{hx}_{(\Tilde{x}^{a}, \Tilde{y}^{a})}}, {\hat{d}^{hy}_{(\Tilde{x}^{a}, \Tilde{y}^{a})}}) $. The displacement loss is defined as:
 
\vspace{-1mm}
\begin{equation}
\begin{aligned}
L_{ah} = & \frac{1}{N}\sum_{ (\Tilde{x}^{a}, \Tilde{y}^{a})  \in {\Tilde{S}^a}}|{\hat{d}^{hx}_{(\Tilde{x}^{a}, \Tilde{y}^{a})}}- {\Tilde{d}^{hx}_{(\Tilde{x}^{a}, \Tilde{y}^{a})}}|\\
& + |{\hat{d}^{hx}_{(\Tilde{x}^{a}, \Tilde{y}^{a})}}-  {\Tilde{d}^{hy}_{(\Tilde{x}^{a}, \Tilde{y}^{a})}}|,
\end{aligned}
\end{equation}
where $\Tilde{S}^a $ denotes the ground-truth interaction point sets in the training set.
$N = | \Tilde{S}^a | $ is  the number of interaction points. 
The loss function for displacement from the interaction point to the object point $L_{ao}$ has the same form. 

\textbf{Triplet matching.}  Two aspects are considered to judge whether a human/object point can be matched with the interaction point. The human/object needs to:   1) be close to the coarse human/object point generated by interaction point plus the displacement and  2) have high confidence scores. On basis of these, for the detected interaction point $(\hat{x}^{a},\hat{y}^{a})$,  we rank the points in the detected human point set  $\hat{S}^{h}$ by Equation \ref{eq:match} and select the optimal one.
\vspace{-2mm}
\begin{equation}
\label{eq:match}
\begin{aligned}(\hat{x}_{opt}^{h}, \hat{y}_{opt}^{h})=  & \underset{(\hat{x}^{h}, \hat{y}^{h}) \in \hat{S}^ h}{\mathop{\arg \min} } \frac{1}{C^h_{(\hat{x}^{h},\hat{y}^{h})}} \\ &({|(\hat{x}^{a},\hat{y}^{a})-({\hat{d}^{hx}_{(\hat{x}^{a}, \hat{y}^{a})}}, {\hat{d}^{hy}_{(\hat{x}^{a}, \hat{y}^{a})}})-(\hat{x}^{h}, \hat{y}^{h})}|  ) \end{aligned},
\end{equation}
where  $C^h_{(\hat{x}^{h},\hat{y}^{h})}$ denotes the confidence score for human point $(\hat{x}^{h},\hat{y}^{h})$. The optimal object box 
$(\hat{x}_{opt}^{o}, \hat{y}_{opt}^{o})$ can be calculated similarly.

\subsection{Loss and Inference}
The final loss can be obtained by weighted summing the above losses:
\vspace{-1mm}
\begin{equation}
    L = L_a + L_h + L_o +  \lambda(L_{ah} + L_{ao} + L_{wh}) + L_{off}
\end{equation}
where we set the $\lambda$ as $0.1$ following 
\cite{law2018cornernet,zhou2019objects}.
$L_a$, $L_h$ and $L_o$ are point location loss, $L_{ah}$ and $L_{oh}$ are displacement loss while  $L_{wh}$ and $L_{off}$ are size and offset lose .

During the inference, we firstly do a $3\times3$ max-pooling operation with stride 1 on the predicted human, object and interaction points heatmap, which plays a similar role as NMS. Secondly, we select top $K$ human points $\hat{S}^{h}$, object center points $\hat{S}^{o}$ and interaction points $\hat{S}^{a}$ through the corresponding confidence scores $\hat{C}^{h}$, $\hat{C}^{o}$ and $\hat{C}^{a}$ across all categories. Then, we find the subject point and object point for each selected interaction point by  Equation \ref{eq:match}. 
For each matched human point  $(\hat{x}^h_{opt}, \hat{y}^h_{opt})$, we get the final box as:
\vspace{-1mm}
\begin{equation}
\begin{aligned}
(\hat{x}^{h}_{ref}-\frac{ \hat{w}_{(\hat{x}^{h}_{opt}, \hat{y}^{h}_{opt})}} {2} , \hat{y}_{ref}^{h}-\frac{ \hat{h}_{(\hat{x}^{h}_{opt}, \hat{y}^{h}_{opt})}} {2},  \\
\hat{x}_{ref}^{h} +\frac{ \hat{w}_{(\hat{x}^{h}_{opt}, \hat{y}^{h}_{opt})}} {2}, 
\hat{y}_{ref}^{h}  +\frac{ \hat{h}_{(\hat{x}^{h}_{opt}, \hat{y}^{h}_{opt})}} {2}. 
\end{aligned}
\end{equation}
where $ \hat{x}^{h}_{ref} = \hat{x}^{h}_{opt}+ \hat{\delta}^x_{(\hat{x}^{h}_{opt}, \hat{y}^{h}_{opt})}$ and $\hat{y}_{ref}^{h} = \hat{y}_{opt}^{h}+ \hat{\delta}^y_{(\hat{x}^{h}_{opt}, \hat{y}^{h}_{opt})} $ are the refined location of the human center point.
 $(\frac{ \hat{w}_{(\hat{x}^{h}_{opt}, \hat{y}^{h}_{opt})}} {2}, \frac{ \hat{h}_{(\hat{x}^{h}_{opt}, \hat{y}^{h}_{opt})}} {2})$ is the size of box in the corresponding position. The final HOI detection results are a set of triplets, and the confidence score for the  triplet is $\hat{C}^p_{\hat{x}^{h}_{ref} \hat{y}^{h}_{ref}} \hat{C}^o_{\hat{x}^{o}_{ref} \hat{y}^{o}_{ref}} \hat{C}^a_{\hat{x}^{a}_{ref} \hat{y}^{a}_{ref}}$.

\section{HOI-A Dataset}
The existing datasets such as HICO-Det~\cite{qi2018learning} and V-COCO~\cite{gupta2015visual} have greatly boosted the development of related research. 
However, in practical application, there are limited frequent HOI categories that need to be paid special attention to, which are not emphasized in previous datasets. We then introduce a new dataset called Human-Object Interaction for Application (HOI-A).
\begin{figure}[htb]
\vspace{-2mm}
  \centering
  \includegraphics[width=0.95\linewidth]{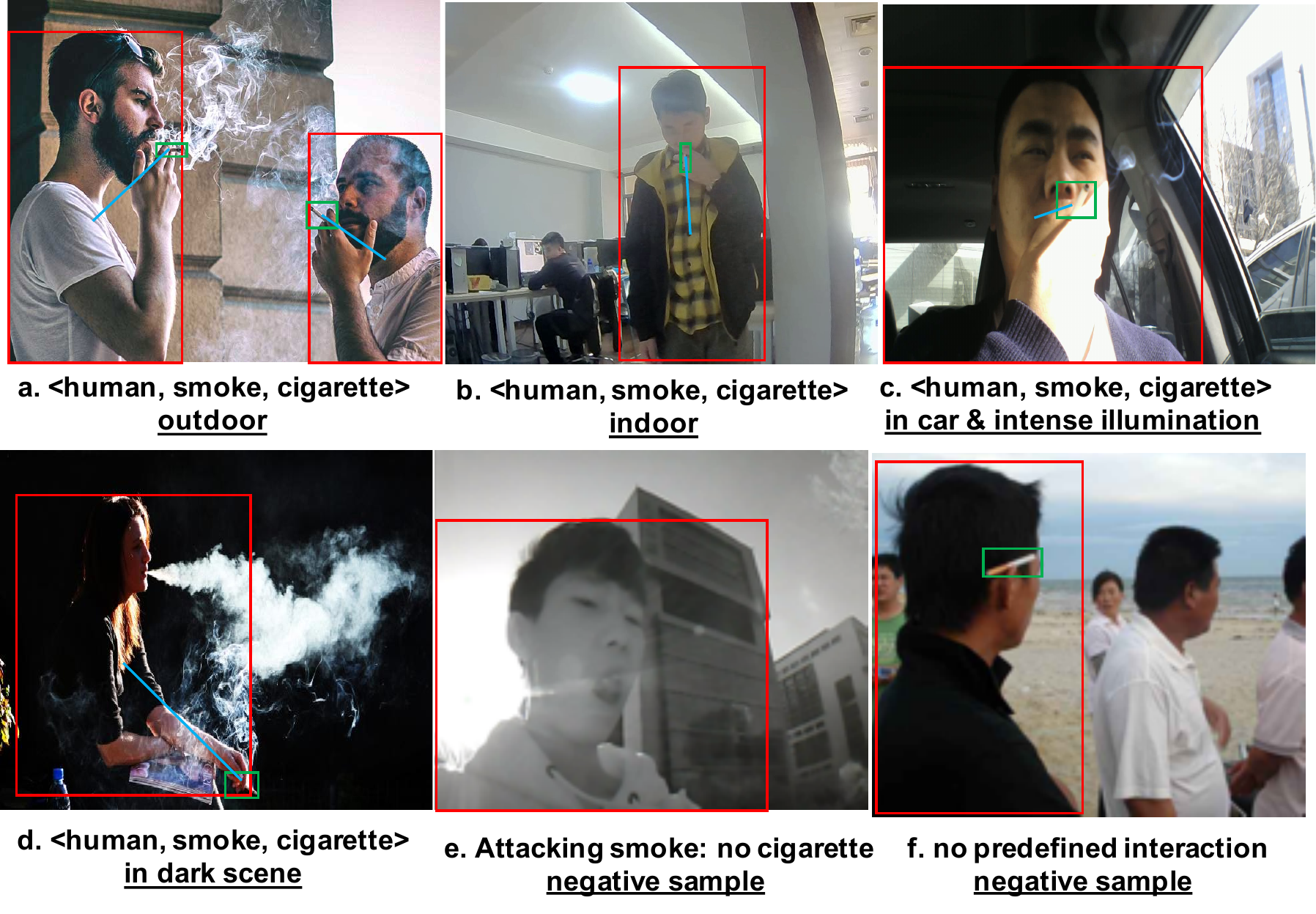}
  \vspace{-1mm}
  \caption{Example images of our HOI-A dataset. We take $<$human, smoke, cigarette$>$ as an example. The (a)-(d) show huge intra-class variations of $<$human, smoke, cigarette$>$ in the wild. The (e)-(f) show two kinds of negative samples. }
  \label{hoiw_view}
    \vspace{-2mm}
\end{figure}


As shown in Table~\ref{tb:verb_def}, we select the categories of verb driven by practical application. Each kind of verb in HOI-A dataset has its corresponding application scenario, for example `talk on' can be applied in dangerous action detection, e.g., if the human is talking on phone in-car, it can be considered as a dangerous driving action.

 \begin{table}[htb!]
 \vspace{-2mm}
\begin{center}

\scriptsize
\begin{tabular}{|c|c|c|}
\hline
Verbs     & Objects                                                                                                                                                                                              & \# Instance \\ \hline
smoking         & cigarette    &  8624   \\ 
talk on             &mobile phone& 18763  \\ 
play (mobile phone) &mobile phone&6728   \\ 
eat                 &food& 831    \\ 
drink               &drink&  6898   \\ 
ride                &bike, motorbike, horse&7111   \\ 
hold                &\begin{tabular}[c]{@{}c@{}}cigarette, mobile phone, food \\ drink, document, computer \end{tabular}&44568  \\ 
kick                &sports ball&365    \\ 
read                &document& 869    \\
play (computer)     &computer& 1402   \\ \hline
\end{tabular}
    
\end{center}
\vspace{-2mm}
\caption{The list and occurrence numbers of the verbs of the corresponding objects in HOI-A dataset.}
\label{tb:verb_def}
\vspace{-2mm}
\end{table}

\vspace{-2mm}
\subsection{HOI-A Construction}
\vspace{-2mm}
We describe the image collection and annotation process for constructing the HOI-A dataset. The first step is collecting candidate images, which can be divided into two parts, namely positive and negative images collections.

\noindent
\textbf{Positive Images Collection.}
We collect positive images in two ways, i.e.,  camera shooting and crawling. Camera shooting is an important way to enlarge the intra-class variations of the data. We employ 50 performers and require them to perform all predefined actions in different scenes and illumination, with various poses, and take photos of them respectively with an RGB camera and an IR camera. 
For data crawling from the Internet, we generate a series of keywords based on the HOI triplet $<$ person, action name, object name$>$, action pair $<$action name, object name$>$ and action name, and retrieve images from the Internet.

\noindent
\textbf{Negative Images Collection.}
Negative Images Collection. There are two kinds of negative samples of the predefined $<$ human,
interaction, object $>$. 
1) The concerned object appears in the image, but the concerned action does not happen. For example, in Figure~\ref{hoiw_view}(f), although the cigarette appears in the image, it is not smoked by a human. Therefore, the image is still a negative sample.
2) Other action similar to the concerned action happens but the concerned object is missing. For example, in Figure~\ref{hoiw_view}(e), the man is smoking at a glance. But a closer look shows there is no cigarette in the image. 
We collect this kind of negative sample in the `attack’ manner. We firstly train a multi-label action classifier based on the annotated positive images. The classifier takes an image as input and outputs the probability of action classification. Then, we let actors perform arbitrarily to attack the classifier without any interacted objects. If the attack is successful, we record this image as a hard negative sample.

\noindent\textbf{Annotation.}
The process of annotation contains two steps: box annotation and interaction annotation. First, all objects in the pre-defined categories are annotated with a box and the corresponding category.
Second, we visualize the boxes in the images with their id and annotate whether a person has the defined interactions with a object. 
The annotator should record the triplet $<$person ID, action ID, object ID$>$. For more accurate annotation, each image is annotated by 3 annotators. The annotation of an image is regarded qualified if at least 2 annotators share the same annotation.

\vspace{-1mm}
\subsection{Dataset Properties}
\vspace{-1mm}
\noindent\textbf{Scale.} 
Our HOI-A dataset consists of 38,668 annotated images, $11$ kinds of objects and $10$ action categories. 
In detail, it contains $43,820$ human instances, $60,438$ object instances and $96,160$ interaction instances. 
There are on average $2.2$ interactions performed per person.  Table~\ref{tb:verb_def} lists the number of instances for each verb which occurs at least $360$ times.  $60\%$ verbs appear more than $6,500$ times. 
To our knowledge, this is already the largest HOI dataset, in terms of the number of images per interaction category. 



\noindent\textbf{Intra-Class Variation.} To enlarge the intra-class variation of the data, each type of verbs in our HOI-A dataset will be captured with three general scenes including indoor, outdoor and in-car, three lighting conditions including dark, natural and intense, various human poses and different angles. Additionally, we shoot the images with two kinds of cameras: RGB and IR.

\section{Experiments}
\begin{table*}[htb!]
  \vspace{-2mm}
  \begin{center}
  
  \small

  \begin{tabular}{cc|ccc|ccc|ccc}
    \hline
    &&\multicolumn{3}{c|}{Default} & \multicolumn{3}{c|}{Know Object} &&\\
  Method              &Feature    & Full & Rare & Non-Rare & Full  & Rare & Non-Rare & Inference Time (ms)  $\downarrow$ & FPS $\uparrow$\\
  \hline
Shen~\emph{et. al}~\cite{shen2018scaling} & A + P &6.46 &4.24 &7.12 & -  & -& -  & -& -\\
  HO-RCNN~\cite{chao2018learning}       &  A + S           & 7.81         & 5.37         & 8.54 & 10.41& 8.94& 10.85 & - &-         \\
  InteractNet~\cite{gkioxari2018detecting}  & A         & 9.94         & 7.16         & 10.77   &- & -&-  & 145 &6.90     \\
  GPNN~\cite{qi2018learning}     & A              & 13.11        & 9.34         & 14.23&- & -&-& 197 +  48 = 245 & 4.08\\
  Xu~\emph{et. al}~\cite{Xu_2019_CVPR} & A + L &14.70 & 13.26& 15.13&- & -&-& - & -\\
  iCAN~\cite{gao2018ican}      & A + S             & 14.84        & 10.45        & 16.15  &16.26 &11.33& 17.73 &92 +  112 = 204 &4.90    \\
  PMFNet-Base~\cite{Wan_2019_ICCV} &A + S &  14.92 &11.42 & 15.96 & 18.83& 15.30& 19.89&- & - \\ 
  Wang~\emph{et. al}~\cite{Wang_2019_ICCV} & A & 16.24 &11.16 &17.75 &17.73 &12.78 &19.21& - & - \\
  No-Frills~\cite{gupta2018no} &A + S + P  & 17.18 & 12.17 & 18.68&- & -&-& 197 + 230 + 67 = 494 & 2.02\\
  TIN~\cite{li2018transferable}      & A + S + P         & 17.22        & 13.51        & 18.32&19.38&	15.38&	20.57&92 + 98 + 323 = 513 &1.95         \\
  RPNN~\cite{Zhou_2019_ICCV} &A + P & 17.35 &12.78 &18.71&-&-&-& - & - \\
  PMFNet~\cite{Wan_2019_ICCV} &A + S + P & 17.46 & \textbf{15.65} & 18.00 &20.34& \textbf{17.47}& 21.20&92 + 98 + 63 = 253 & 3.95 \\ \hline
  PPDM-DLA & A &20.29&	13.06&	22.45&	23.09&	16.14&	25.17& \textbf{27} & \textbf{37.03} \\ 
  PPDM-Hourglass & A & \textbf{21.73}&	13.78&	\textbf{24.10}&	\textbf{24.58}&	16.65&	\textbf{26.84}& 71 & 14.08 \\
  
 \hline          
  \end{tabular}
  \end{center}
  \vspace{-2mm}
    \caption{Performance comparison on the HICO-DET test set. The `A', `P', `S', `L' represent the appearance feature, human pose information,  the spatial feature, and the language feature, respectively.}
  \label{tb:hico}
  \vspace{-4mm}
  \end{table*}
\subsection{Experimental Setting}
\noindent\textbf{Datasets.} To verify the effectiveness of our PPDM, we conduct experiments not only on our HOI-A dataset but also on the general HOI detection dataset HICO-Det~\cite{chao2018learning}. HICO-Det is a large-scale dataset for common HOI detection. It has $47,776$ images ($38,118$ for training and $9,658$ for test), annotated with $117$ verbs including `no-interaction' and $80$ object categories. The $117$ verbs and $80$ objects form $600$ kinds of HOI triplets, where $138$ types of HOIs which appear $\textless10$ times are considered as the rare set, and the rest $462$ kinds of HOIs form the non-rare set.

\noindent\textbf{Metric.} Following the standard-setting in HOI detection task, we use mean average precious (mAP) as the metric. 
If a predicted triplet is considered as a true positive sample, it needs to match a certain ground-truth triplet. Specifically, they have the same HOI class and their human and object boxes have overlap with IOUs large than $0.5$.  There is a slight difference when computing AP on the two datasets. We compute AP per HOI class in HICO-Det and compute AP per verb class in HOI-A dataset.

\noindent\textbf{Implementation Details.}
We use two common heatmap prediction networks as our feature extractor, Hourglass-104~\cite{newell2016stacked,law2018cornernet} and DLA-34~\cite{yu2018deep,zhou2019objects}. Hourglass-104 is a general heatmap prediction network commonly used in keypoint detection and object detection. In PPDM, we use the modified version Hourglass-104 proposed in~\cite{law2018cornernet}. The DLA-34 is a lightweight backbone network, and we apply a refined version proposed in~\cite{zhou2019objects}. The receptive field of the network need large enough to cover the subject and the object. Hourglass-104 has a sufficiently large receptive field, while that of DLA-34 cannot cover the region including the human and the object, due to its relatively shallow architecture.  Thus for the DLA-based model, we concatenate the last three level features and apply a graph-based global reasoning module~\cite{chen2019graph} to enlarge the receptive field for the interaction point and displacement prediction. In the global reasoning module, we set the channels of the node and the reduced feature as $48$ and $96$ respectively. For Hourglass-104, we only use the last-level feature for all the following modules. We initialize the feature extractor with the weights pre-trained on COCO~\cite{lin2014microsoft}. Our experiments are all conducted on the Titan Xp GPU and CUDA 9.0.  

During training and inference, the input resolution is $512\times512$ and the output is $128\times128$. PPDM is trained with Adam on 8 GPUs. We set the hyper-parameter following~\cite{zhou2019objects}, which is robust to our framework. We train the model based on DLA-34 with a 128 sized mini-batch for 110 epochs, with a learning rate of 5e-4 decreased to 5e-5 at the 90th epoch. For the hourglass-104 based model, we train it with a batch size of 32 for 110 epochs, with a learning rate of 3.2e-4 decreased by 10 times at the 90th epoch. We follow~\cite{law2018cornernet,zhou2019objects} applying data augmentation, i.e., random scale and random shift to train the model and there is no augmentation during inference. We set the number of selected predictions $K$ as 100.

\vspace{-1.5mm}
\subsection{Comparison to State-of-the-art}
\begin{figure*}[h]
  \vspace{-2mm}
  \centering
  \includegraphics[width=0.95\linewidth]{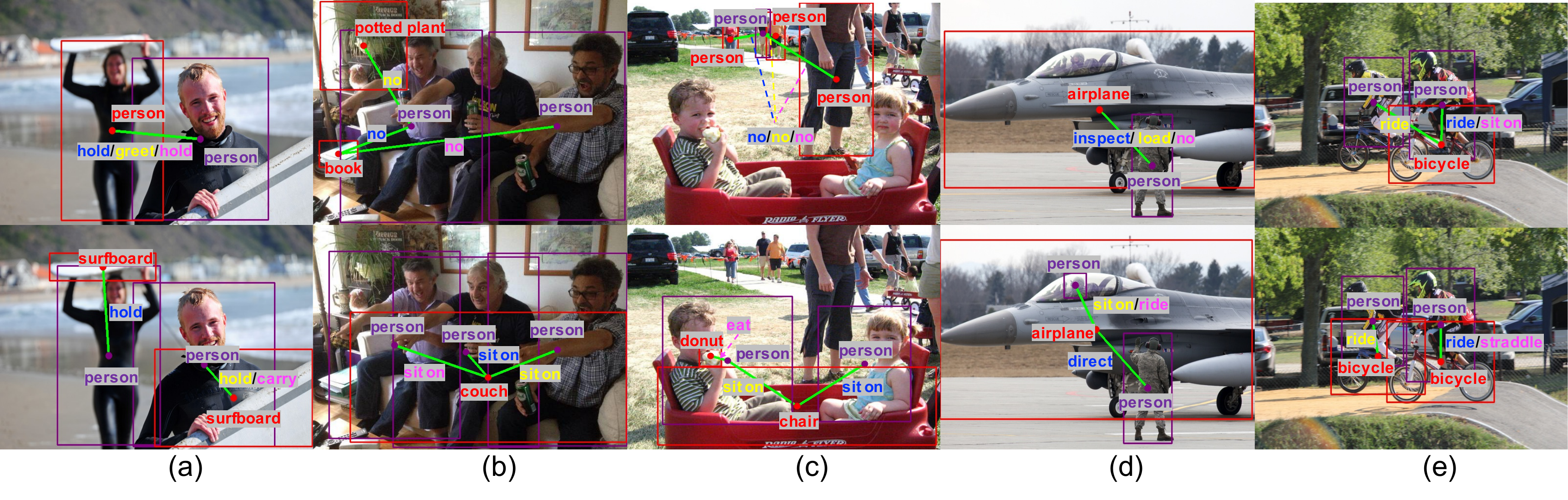}
  \vspace{-2mm}
  \caption{Visualization results compared with iCAN on HICO-Det. The first row is the prediction of iCAN and the second row by PPDM. Purple denotes the subject and red is the object. If a subject has interaction with an object, they will be linked by a green line. We show results with top-3 confidence per image: 1-blue, 2-yellow, 3-pink. The `no' denotes `no-interaction'.}
  \label{fig:vis}
  \vspace{-4mm}
\end{figure*}

\vspace{-1.5mm}
We compare PPDM with state-of-the-art methods on two datasets. The quantitative results can be seen in Table~\ref{tb:hico} and Table~\ref{tb:hoiw_exp}, and the qualitative results are presented in Figure~\ref{fig:vis}. The compared methods mainly use a pre-trained Faster R-CNN~\cite{ren2015faster} to generate a set of human-object proposals, which are then fed into a pairwise classification network. As shown in Table~\ref{tb:hico}, to more accurately classify the HOI, many methods use additional human pose feature or language feature. 

\vspace{-2.5mm}
\subsubsection{Quantitative Analysis}
\vspace{-1.5mm}
\noindent\textbf{HICO-Det.} See table~\ref{tb:hico}. Our PPDM-DLA and PPDM-Hourglss both outperform all previous state-of-the-art methods. Specifically, our PPDM-Hourglass achieves a significant performance gain ($24.5\%$) comparing to the previous best method PMFNet~\cite{Wan_2019_ICCV}. We can see the previous methods with mAP greater than $17\%$ all use the human pose as an additional feature, while our PPDM only uses the appearance feature. Performance of PPDM is slightly lower than PMFNet on the rare subset. However, the baseline model in PMFNet without using human pose information only achieves $11.42\%$ mAP on the rare-set. The performance gain on the rare-set may mainly come from the additional human pose feature. The human structural information plays an
important role in understanding human actions, thus we regard how to utilize human context in our framework as a significant future work.
\begin{table}[htb!]
  \vspace{-2mm}
  \begin{center}
  \small
  \begin{tabular}{ccc}
    \hline
  Method           &mAP (\%) & Time (ms)\\
  \hline
  Faster Interaction Net~\cite{picleadboard} &56.93&-\\
  GMVM~\cite{picleadboard} &60.26&-\\
  C-HOI~\cite{Zhou_2020_CVPR} &66.04&-\\
  \hline
  iCAN~\cite{gao2018ican} &44.23& 194 \\
  TIN~\cite{li2018transferable} & 48.64 & 501 \\
  \hline
  PPDM-DLA & 67.45 & \textbf{27}\\
  PPDM-Hourglass & \textbf{71.23} & 71\\
  \hline
  \end{tabular}
  \end{center}
  \vspace{-2mm}
  \caption{ Performance comparison on HOI-A test set.}
  \label{tb:hoiw_exp}
\vspace{-2mm}
 \end{table}

\noindent\textbf{HOI-A.}
The compared methods in HOI-A dataset are composed of two parts. Firstly, we select the top-3 methods from the leaderboard of ICCV 2019 PIC challenge HOI detection track~\cite{picleadboard}, which was held by us based on HOI-A dataset. Comparing to the top-1 method, C-HOI~\cite{Zhou_2020_CVPR}, which uses a very strong detector, our methods still outperform it.  Secondly, we choose two open-source state-of-the-art methods, iCAN~\cite{gao2018ican} and TIN~\cite{li2018transferable}, as the baselines on our HOI-A dataset. We first pre-train Faster R-CNN with FPN and ResNet-50 on HOI-A dataset, and then follow their original settings  to train the HOI classifier. The results show our PPDM outperforms the two methods by a significant margin. Additionally, for our selected interaction types with practical significance, our PPDM can achieve high performance, which is practically applicable.

\vspace{-3mm}
\subsubsection{Qualitative Analysis}
\vspace{-2mm}
We visualize the HOI prediction with the top-3 confidence score on HICO-Det dataset based on PPDM-DLA, and compare our results with the typical two-stage method iCAN~\cite{gao2018ican}. As shown in Figure~\ref{fig:vis}, we select some representative failure cases of the two-stage methods. We can see iCAN tends to focus on the human/object with a high detection score but without interaction. In Figure~\ref{fig:vis}(b) and Figure~\ref{fig:vis}(c), due to the huge imbalance between positive/negative samples,  iCAN easily produces high confidence for the `no-interaction' type. In Figure~\ref{fig:vis}(d), the person sitting on the airplane is so small that it cannot be detected. However, our PPDM can accurately predict the HOI triplets with high confidence in these cases. Because PPDM is not dependent on the proposals. Moreover, PPDM concentrates on the HOI triplets understanding.

\begin{table}[htb!]
\vspace{-2mm}
  \begin{center}
  \footnotesize
  \begin{tabular}{|c|ccccc|}
    \hline
  &Method           &Full & Rare & Non-Rare & Time \\
  \hline
   1&Basic Model& 19.94 &13.01& 22.01&24 \\
  2&+ Feature Fusion& 20.00 &12.56& 22.22& 26\\
  3&+ Global Reasoning& 19.85 &12.99 & 21.90& 26\\
  \hline
  4&Union Center & 18.65 & 12.11 &20.61& 27 \\ \hline
  5& PPDM-DLA & 20.29 &13.06 & 22.45 & 27 \\ \hline
  
  \end{tabular}
  \end{center}
  \vspace{-2mm}
  \caption{Component analysis on HICO-Det Test Set.}
  \label{tb:abla}
    \vspace{-5mm}
 \end{table}

\vspace{-3mm}
\subsubsection{Efficiency Analysis}
\vspace{-2mm}
We compare the inference speed on a single Titan Xp GPU with the methods which have released code or reported the speed. As shown in Table~\ref{tb:hico}, PPDM with DLA and Hourglass are both faster than other methods by a large margin. PPDM-DLA is the only real-time method, which only takes $27$ms for inference. Concretely, the inference time of two-stage HOI detection methods can be divided into proposal generation time and  HOI classifier time.
Besides, the pose based methods take extra time to estimate human key-points. It can be seen that the speed of PPDM-DLA is faster than any stage of the compared methods.

\vspace{-3mm}
\subsection{Component Analysis}
\vspace{-2mm}

We analyze the proposed components in PPDM from quantitative and  qualitative views.

\begin{figure}[htb]
  \vspace{-2mm}
  \begin{center}
  \includegraphics[width=0.95\linewidth]{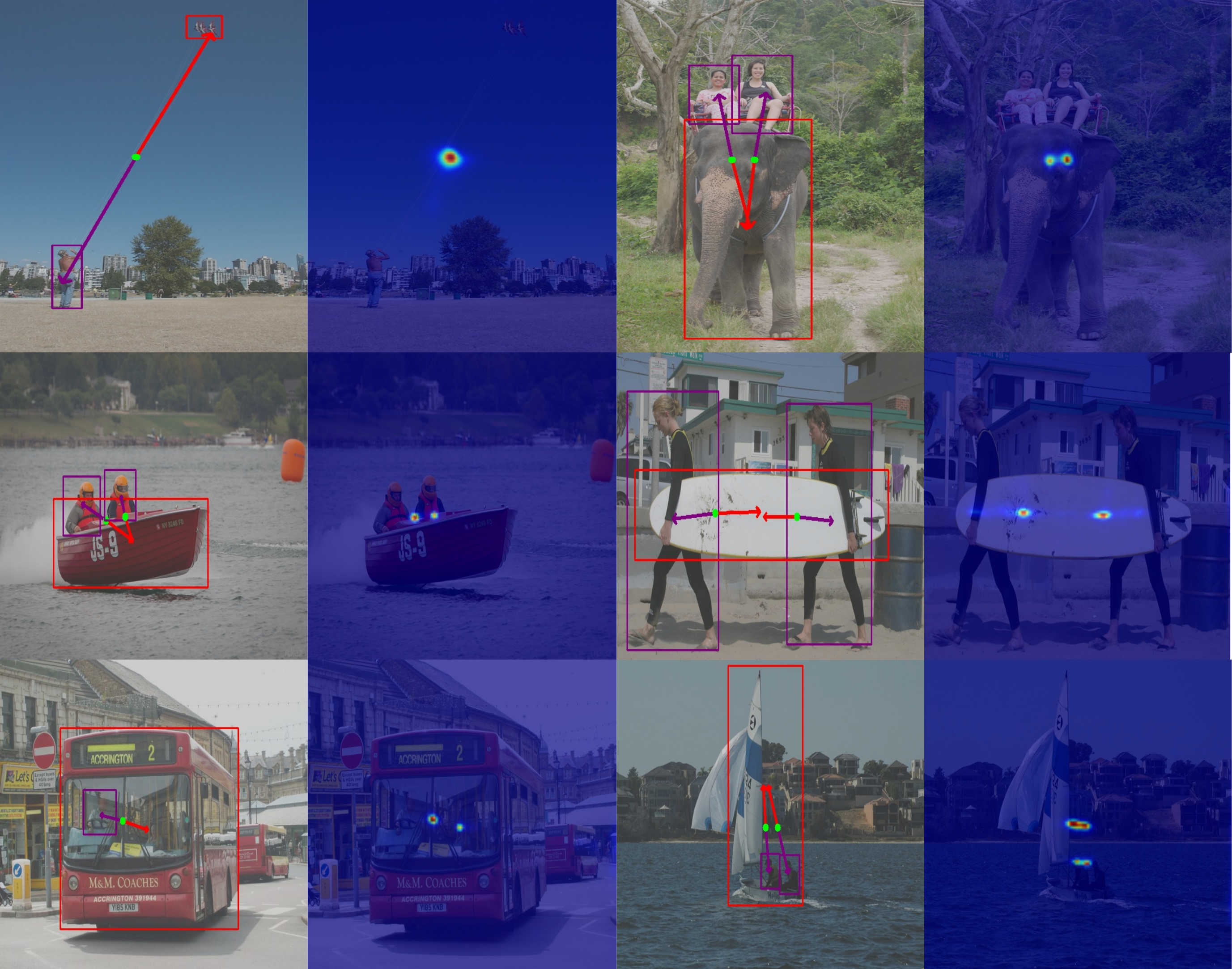}
 \end{center}
 \vspace{-2mm}
  \caption{Visualization of interaction points heatmaps and  displacements. Red and purple line represent displacements from interaction point (green) to human and object. }
  \label{fig:heatmap}
  \vspace{-3.5mm}
\end{figure}

\noindent\textbf{Feature Extractor.} We analyze effectiveness of the additional modules in DLA backbone, i.e., feature fusion and global reasoning. The first row in Table~\ref{tb:abla} represents the basic framework with DLA, where we predict the interaction only based on the last-level feature. It shows that the basic model can still outperform all existing methods. It proves the effectiveness of our designed framework. The second row  and third row show the results of the basic model with the feature fusion and a global reasoning module respectively, it can be seen in Table~\ref{tb:abla} that the performance only change little. If we add a these two settings to the basic framework as the same time,  that the performance improves by $0.35\%$ mAP. We conclude that a larger receptive field and global context are helpful to interaction prediction.

\noindent\textbf{Point Detection.}
To verify whether the midpoint of two center points is the best choice to predict the interaction, we perform an experiment based on the interaction point at the center of the union of human and object boxes, which is another suitable location to predict the interaction. See the 4th row in Table~\ref{tb:abla}. The mAPs drop 1.64 point compared with PPDM-DLA. It is common that two objects interact with the same person and may locate in the human box, in which case the center points of their union boxes overlap. Additionally, we analyze our interaction point qualitatively. As shown in Figure~\ref{fig:heatmap}, the predicted interaction almost accurately locates at the midpoint of the human/object points, though the human is apart from the object or in the object.

\noindent\textbf{Point Matching.}
To further understand the displacement, we visualize the displacements in Figure~\ref{fig:heatmap}. We can see the interaction point plus the corresponding displacement is very close to the center point of the human/object box, even though the human/object is hard to be detected.

\vspace{-3mm}
\section{Conclusion}
\vspace{-2mm}
In this paper, we propose a novel one-stage framework and a new dataset for HOI detection. Our proposed method can outperform the existing methods by a margin also with a significantly faster speed. It breaks the limits of the traditional two-stage methods and directly predicts the HOI by a parallel framework. Our proposed HOI-A dataset is more inclined to practical application for HOI detection. For future work, we plan to explore how to utilize human context in our framework. Additionally, we plan to enrich the action categories for HOI-A dataset.

{\small
\bibliographystyle{ieee_fullname}
\bibliography{egbib}
}

\end{document}